\documentclass[10pt,twocolumn,letterpaper]{article}

\usepackage[ruled,linesnumbered]{algorithm2e}
\usepackage{amsmath}
\usepackage{colortbl}
\usepackage{multirow}
\usepackage{iccv}              
\definecolor{iccvblue}{rgb}{0.21,0.49,0.74}
\usepackage[pagebackref,breaklinks,colorlinks,allcolors=iccvblue]{hyperref}

\title{Language-Driven Dual Style Mixing for Single-Domain Generalized\\Object Detection}

\author{
Hongda Qin$^{1,}$\thanks{Equal contribution} \quad Xiao Lu$^{2,*}$ \quad Zhiyong Wei$^{1}$ \quad Yihong Cao$^{3}$ \quad  Kailun Yang$^{3}$ \quad Ningjiang Chen$^{1,}$\thanks{Correspondence: chnj@gxu.edu.cn}\\
$^{1}$Guangxi University \quad $^{2}$Hunan Normal University \quad $^{3}$Hunan University\\
}

\begin{document}
\maketitle

\begin{abstract}
Generalizing an object detector trained on a single domain to multiple unseen domains is a challenging task. Existing methods typically introduce image or feature augmentation to diversify the source domain to raise the robustness of the detector. Vision-Language Model (VLM)-based augmentation techniques have been proven to be effective, but they require that the detector's backbone has the same structure as the image encoder of VLM, limiting the detector framework selection. To address this problem, we propose \textbf{L}anguage-\textbf{D}riven \textbf{D}ual \textbf{S}tyle Mixing (LDDS) for single-domain generalization, which diversifies the source domain by fully utilizing the semantic information of the VLM. Specifically, we first construct prompts to transfer style semantics embedded in the VLM to an image translation network. This facilitates the generation of style diversified images with explicit semantic information. Then, we propose image-level style mixing between the diversified images and source domain images. This effectively mines the semantic information for image augmentation without relying on specific augmentation selections. Finally, we propose feature-level style mixing in a double-pipeline manner, allowing feature augmentation to be model-agnostic and can work seamlessly with the mainstream detector frameworks, including the one-stage, two-stage, and transformer-based detectors. Extensive experiments demonstrate the effectiveness of our approach across various benchmark datasets, including real to cartoon and normal to adverse weather tasks. The source code and pre-trained models will be publicly available at \url{https://github.com/qinhongda8/LDDS}.
\end{abstract}    
\section{Introduction}
\label{sec:intro}

Due to the reliable performance of existing fully-supervised object detection methods~\cite{redmon2016you,ren2015faster} in recognizing and localizing objects, they have been widely applied in fields such as autonomous driving, drones, and intelligent robots. However, in real-world applications, object detectors face a challenge where a domain distribution gap between the training data and the target deployment scenarios, \eg, adverse weather~\cite{sakaridis2018semantic,wu2021VDD} or out-of-distribution style~\cite{inoue2018cross,johnson2016driving}, leading to unexpected degradation in detection performance.

\begin{figure}[t]
  \centering
   \includegraphics[width=1.0\linewidth]{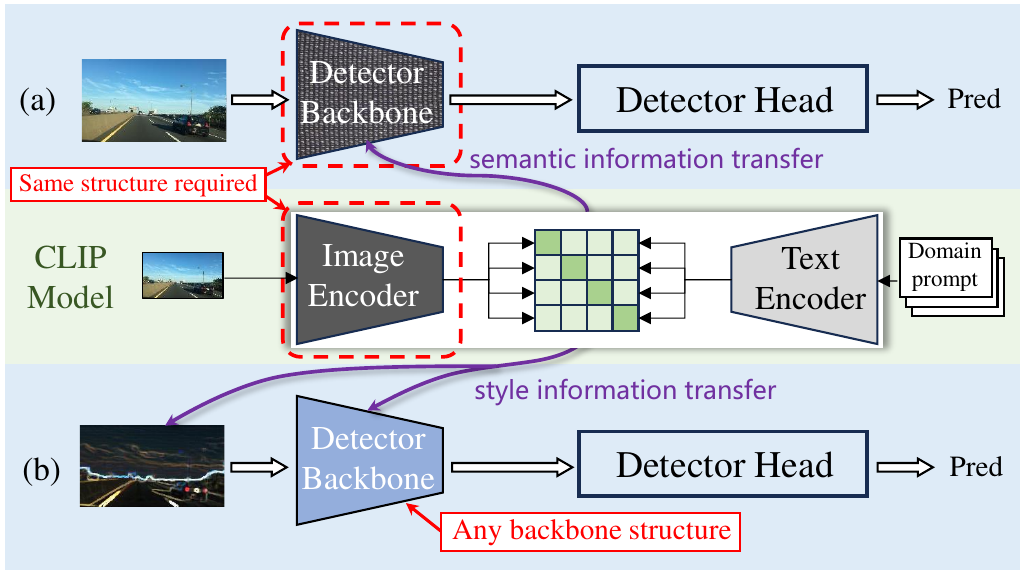}

    \caption{Existing VLM-based methods~\cite{vidit2023clip,fahes2023poda} (a) require that the detector's backbone has the same architecture as the image encoder, resulting in the limitation of detector framework selection. Besides, they only perform feature augmentation of the detector’s backbone, can not effectively transfer the semantic information embedded in the VLM. Our method (b) avoids these limitations with proposed dual-style mixing schemes for both image and feature augmentation.}
   \label{fig:abstract}
\end{figure}

To tackle this challenge with minimal cost, few-shot domain adaptation~\cite{gao2023asyfod, Few-shotCross-domain, wang2019few, zhong2022pica_few, gao2022acrofod_few} adjusts the detector by generating numerous samples using a small amount of labeled target domain images. Unsupervised domain adaptation~\cite{chen2018domain,he2019multi,chen2020harmonizing,deng2021unbiased,chen2022learning,kennerley2024cat,do2024d3t}, on the other hand, reduces domain gaps by leveraging unlabeled target domain images through techniques like feature alignment or self-training. While these methods transfer knowledge to target domains at low resource costs, they still rely on pre-collecting target domain data. In real-world scenarios, collecting target domain images in advance is daunting, such as in rare adverse weather like hailstorms or sandstorms. Consequently, Single-Domain Generalization (SDG)~\cite{fan2021adversarially,chen2023meta,cugu2022attention,xu2023simde,chen2023center,wang2021learning} has gained attention, with its aim to generalize to unseen domains using a single model trained on a source domain.

Existing SDG object detection methods~\cite{wu2022single,vidit2023clip,
danish2024improving,fahes2023poda,qi2024doubleaug} primarily focus on image or feature augmentation. Some image augmentation methods~\cite{danish2024improving,qi2024doubleaug} corrupt source images and align feature maps between the original and corrupted images. These methods require careful selection of augmentation types in the early stage, adding extra cost and limiting performance gain to generalize to unseen domains. Fahes~\etal~\cite{fahes2023poda} try to use Vision-Language Models (VLMs), \eg, CLIP~\cite{radford2021learning}, to drive  style transfer models for image augmentation~\cite{kwon2022clipstyler}. They find that these models, like CLIPstyler~\cite{kwon2022clipstyler}, often introduce blurring and distortion to object details, leading to negative effects. Other feature augmentation methods~\cite{vidit2023clip,fahes2023poda} transfer semantic information in latent space to the detector's backbone by fine-tuning the pre-trained VLMs. They shed new light on exploring multi-modal learning scheme for SDG, avoiding the careful selection of augmentation. However, as shown in~\cref{fig:abstract}, these methods require that the detector's backbone has the same structure as the VLM's image encoder to ensure the transfer of semantic information. This limits the detector framework selection especially in scenarios with high real-time requirements, since the image encoder of VLM is typically built on large networks, \eg, ResNet-101~\cite{he2016deep} or ViT-B/32~\cite{dosovitskiy2020image}. Besides, these methods only perform feature augmentation of the detector’s backbone, can not fully transfer the semantic information in the VLM.

Given the issues mentioned above, we naturally consider the following challenge. \textit{How to effectively transfer the semantic information of VLM to the detector while avoiding the structural restrictions on the backbone?} In response, the motivation of our work can be outlined in two aspects. 1) Extract the latent semantic features from the VLM and make them explicit, instead of only fine-tuning the intermediate features. 2) Adapt the extracted semantic information for the image and feature augmentation to the detector, free from the limitations of specific network structure.

Based on this, we propose a simple method called Language-Driven Dual Style Mixing (LDDS) for SDG object detection. Specifically, we first construct style prompts and leverage VLM to transfer the explicit style semantic information to the image generation network, to produce style diversified images. Then we propose the dual style mixing method for image and feature augmentation in SDG object detection, which involves two steps. 1) During the image augmentation phase, we capture the global style information, \eg, brightness and contrast, from both the source and the diversified images, and then design an image-level style mixing to augment the source images. 2) For the feature augmentation phase, we introduce a double-pipeline processing scheme for the feature-level style mixing and the source feature augmentation, respectively. These two steps allow us to preserve the integrity of the original object representation in the images while incorporating the style information of VLM. In addition, to mitigate potential training instability caused by double rounds of style mixing, we design a smoothing strategy for the feature-level style mixing stage.

Our contributions can be summarized as follows.
\begin{itemize}
\item We propose a language-driven dual style mixing framework for SDG object detection, which transfers the semantic information from VLMs to enhance the detector's performance in unseen scenarios. Our framework is model-agnostic and can be seamlessly integrated into object detectors without structure limitations.

\item We propose a dual style mixing method that diversifies source domains through both image-level and feature-level to better utilize the style information guided by language for semantic augmentation. Moreover, we propose a smoothing strategy in the feature-level style mixing stage to avoid the conflicts that may be caused by the double rounds of style mixing.

\item We evaluate our method on a wide range of SDG object detection benchmarks, including the real to cartoon and normal to adverse weather tasks, with the mainstream object detector frameworks, including one-stage, two-stage, and transformer-based detectors. All the experimental results demonstrate the effectiveness of our method.
\end{itemize}

\begin{figure*}[ht]
\centering{\includegraphics[width=0.98\textwidth]{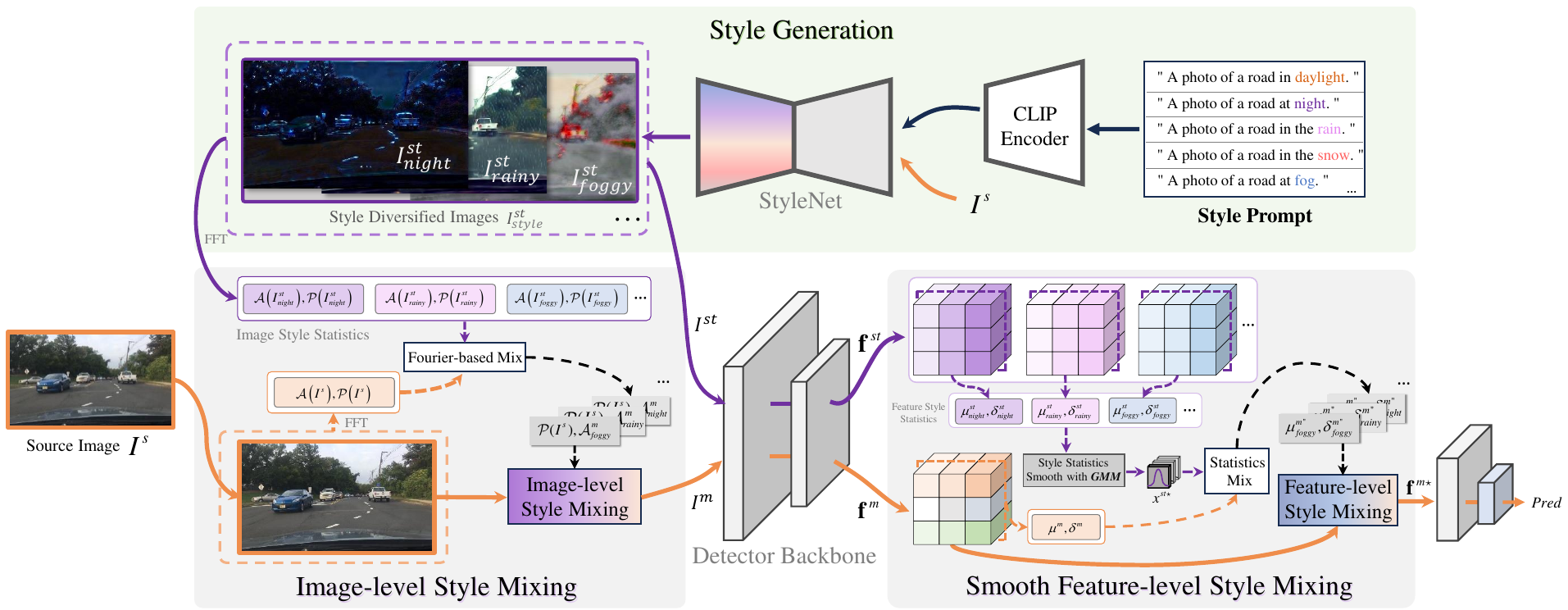}}
\caption{The pipeline of LDDS, which consists of style generation, image-level and smooth feature-level style mixing. We first feed the textual embeddings of the style prompts obtained from the CLIP together with the source image $I^s$ into the StyleNet to generate style diversified images ${I}^{st}$. Then, the source image ${I}^{s}$ and ${I}^{st}$ undergo image-level style mixing in the frequency domain to produce a style mixed image $I^{m}$. We perform a style statistics smooth operation on the statistical style data of $\mathbf{f}^{st}$ to avoid the possible style conflicts. Finally, the feature-level style mixing is performed with the statistical style data of the source image feature map $\mathbf{f}^m$. The resulting dual mixed style feature map $\mathbf{f}^{m\star}$ serves as fully supervised data for the detector.}
\label{fig3}
\end{figure*}

\section{Related Work}
\label{sec:related}

\noindent\textbf{Object detection.} Object detection is a key foundational task in the field of computer vision. Existing mainstream detectors can be typically divided into one-stage, \eg, SSD~\cite{liu2016ssd}, FCOS~\cite{tian2020fcos}, and the YOLO series~\cite{redmon2016you,redmon2017yolo9000,bochkovskiy2020yolov4,yolov8}, two-stage, \eg, Faster R-CNN~\cite{ren2015faster} and Mask R-CNN~\cite{he2017mask}, and transformer-based frameworks, \eg, DETR~\cite{carion2020end} and RT-DETR~\cite{zhao2024detrs}. In real-world applications, the choice of detectors with different frameworks depends on the balance between precision and inference speed.

\noindent\textbf{Single-domain generalized object detection.} Several works~\cite{wu2022single,danish2024improving,qi2024doubleaug,vidit2023clip,fahes2023poda} have addressed SDG object detection problem to allow models trained in a single source domain to generalize across multiple target domains. Wu~\etal~\cite{wu2022single} propose to employ cyclic disentangled self-distillation to extract domain-invariant representations while separating domain-specific features. Danish~\etal~\cite{danish2024improving} explored a range of image augmentation techniques and align multiple views to allow the detector to unify source diversity with detection alignment. Qi~\etal~\cite{qi2024doubleaug} achieved data diversification by applying color perturbations and employing style feature augmentation. Recently, approaches use VLMs to introduce richer semantic information augmentation via textual prompts. Vidit~\etal~\cite{vidit2023clip} utilized CLIP to embed the source domain images into prompt-specified domains, and then trained detectors with optimized semantic augmentation. Fahes~\etal~\cite{fahes2023poda} use prompts to guide the image encoder to capture diverse low-level feature statistics for source features augmentation. While these methods achieve performance close to supervised methods by utilizing latent information of VLM, they require the detector's backbone to be the same with the image encoder of the VLM, limiting the selection of detector framework in real-world scenarios. Therefore, we further consider methods agnostic to the detector architecture to capture the latent semantic information of VLMs.

\noindent\textbf{Data mixing augmentation.} Data augmentation that mixes multiple samples into a single one has been proven to be effective in preventing model over-fitting~\cite{verma2019manifold,yun2019cutmix,chen2022transmix,kim2020puzzle,yao2022c,islam2024diffusemix}. Based on this, data mixing has been successfully applied in domain adaptation~\cite{mattolin2023confmix,zhu2023patch,tranheden2021dacs,hu2023mixed} and domain generalization~\cite{hong2021stylemix,zhou2024mixstyle}. However, in SDG object detection, direct data mixing offers limited augmentation benefits and may introduce potential detail degradation, restricting generalization to broader domains. 
In this work, we aim to use a language-guided style generation approach to perform information-enriched and detail-preserving data mixing.

\section{Method}

\textbf{Problem Setting.} We define the labeled data as the source domain $D_s = \left\{({I}_i^s,{b}_i^s)\right\}_{i = 1}^{N_s}$, with ${b}_i^s$ representing the bounding box and category for the $i$-th image ${I}_i^s$ among $N_s$ images. Generally, the source data consists of images from a single specific domain, \eg, sunny traffic scenes or cartoon illustrations. The unseen domain is defined as the target domain $D_t$, where both the images and annotations are unavailable during training and span a variety of different scenes. The key to this problem is how to effectively use the source domain data to improve the generalization of the object detector to unseen domains.

The framework of our proposed LDDS for SDG object detection is shown in~\cref{fig3}. In the style generation stage, our aim is to transfer the style semantics information embedded in the VLM to the image generation network, to generate style diversified images. To do that, we feed the textual style embeddings obtained from the CLIP and the source image $I^s$ to StyleNet for diversified images generation. Then the generated images $I^{st}$ and the source images are mixed in frequency domain in the image-level style mixing stage. With these two stages, we can obtain the language-driven global style semantic augmented image $I^{m}$. In the feature-level style mixing stage, the feature style statistics of $I^{st}$ are further processed by a Gaussian Mixture Model (GMM) to obtain the smoothed feature statistics ${x^{st\star}}$. This operation can avoid the style conflict that may be caused by the double-round mixing. After that, the statistics of $\mathbf{f}^m$ are mixed with ${x^{st\star}}$ for local semantic style mixing. The resulting dual
mixed style feature map $\mathbf{f}^{m*}$ serves as fully supervised data for the detector.

\subsection{Style Generation}
To exploit the explicit latent semantic information embedded in the VLM to generate style diversified images, we construct a style generation module to achieve this goal. Inspired by the recently proposed CLIPstyler~\cite{kwon2022clipstyler}, we construct several textual prompts to describe diverse domain styles, and the embeddings of which are obtained with the pre-trained CLIP. Then the textual embeddings together with the source image $I^s$ are fed into a pre-trained UNet-based StyleNet~\cite{ronneberger2015u,gatys2016image} to generate style diversified image $I^{st}$. To generate the prompt-consistent images, we fine-tune StyleNet with the loss $\mathcal{L}_s$ following~\cite{kwon2022clipstyler}:
\begin{equation}
\label{eq_clipsty}
{{\cal L}_{s}} = {{\cal L}_{c}}({I}^{st},{I}^s)+{{\cal L}_{d}}(CLIP({I}^{st}),CLIP(text)), 
\end{equation}
where the content loss ${{\cal L}_{c}}$ is to maintain the content information of the input source image ${I}^s$, and the CLIP loss ${{\cal L}_{d}}$ is employed to match the style between the output ${I}^{st}$ and prompt text feature $CLIP(text)$. Based on this, only a short period of fine-tuning is required to achieve diverse style modulation of the source domain images. The second column of~\cref{fig:fft_compare} illustrates the generated samples.

Carefully selected prompts can yield higher-quality style information. To demonstrate the generalization capability of the proposed method, we use unmodified prompts as the driving text input $text$, \eg, the source image prompt is represented by ``a photo'' while the unseen domain prompt is defined as ``a photo of the road in the \underline{rain}''. 
\subsection{Image-level Style Mixing}
Directly putting these images ${I}^{st}$ as training data for SDG
has been proven to be limited in effectiveness~\cite{fahes2023poda}, as they introduce noise and distort object details in the style diversification step. Therefore, our goal is to utilize the style information, not the entire image data. At the image level, after applying the Fourier transform $\mathcal{F}$~\cite{nussbaumer1982fast}, the phase $\mathcal{P}$ retains content information, while the amplitude $\mathcal{A}$ carries image-level style information~\cite{yang2020fda, xu2021fourier}. We convert both source and diversified images into the frequency domain as follows.
\begin{equation}
\label{eq_fft1}
\mathcal{F}({I}^s)(u,v) = \sum\limits_{x = 0}^{M - 1} {\sum\limits_{y = 0}^{N - 1} {I}^s } (x,y){e^{ - 2\pi j\left( {\frac{{ux}}{M} + \frac{{vy}}{N}} \right)}},
\end{equation}
\begin{equation}
\label{eq_fft2}
\mathcal{F}({I}^s)(u,v) = {\mathcal{A}(I^s)}(u,v){e^{j{\mathcal{P}(I^s)} (u,v)}},
\end{equation}
where $M$ and $N$ correspond to the width and height of the image, while 
$u$ and $v$ denote the frequency components in the horizontal and vertical directions, and $j$ is the imaginary unit. By performing similar operations, we acquire the amplitude ${\mathcal{A}(I^{st})}(u,v)$ and phase ${\mathcal{P}(I^{st})}(u,v)$ of the ${I}^{st}$. At this point, we naturally proceed with amplitude mixing to implement image-level style mixing as follows.
\begin{equation}
\label{eq_fft_m}
\mathcal{A}^m_i = (1 - \omega ){\mathcal{A}({I}^s_i)} + \omega {\mathcal{A}({I}^{st}_i)},
\end{equation}
where $\omega$ is randomly sampled from the range $({\gamma}_1,{\gamma}_2)$, and the mixed amplitude $\mathcal{A}^m_i(u,v)$ with the corresponding source image phase ${\mathcal{P}^{s}_i}(u,v)$ forms a new frequency domain data. By applying an inverse Fourier transform ${\mathcal{F}^{-1}}$, we produce the image-level style mixed image ${{I}^m_i} = {\mathcal{F}^{-1}}[\mathcal{A}^m_i(u,v){e^{j{\mathcal{P}_i(I^s)} (u,v)}}]$. This achieves our goal of extracting the style information generated by the VLM at the image-level style mixing.

It should be noted that our implementation differs from the existing amplitude mixing and amplitude transfer approaches~\cite{yang2020fda, xu2021fourier}. 
For amplitude mixing~\cite{yang2020fda}, only the amplitude information within the source images is mixed to achieve amplitude augmentation due to the lack of other style images. 
In contrast, amplitude transfer~\cite{xu2021fourier} relies on existing target domain images to exchange amplitude information to bring source image style closer to the target images. 
Although we obtained style diversified images from the style generation module, they still do not fully match the real data distribution. 
Therefore, for a source domain image ${I}^s_i$, we use its corresponding generated image ${I}^{st}_i$ for style mixing, as shown in the third column in~\cref{fig:fft_compare}, which allows for more effective style augmentation while minimizing negative noise and preserving object details.

\subsection{Smooth Feature-level Style Mixing}
\begin{figure}[t]
  \centering
   \includegraphics[width=1.0\linewidth]{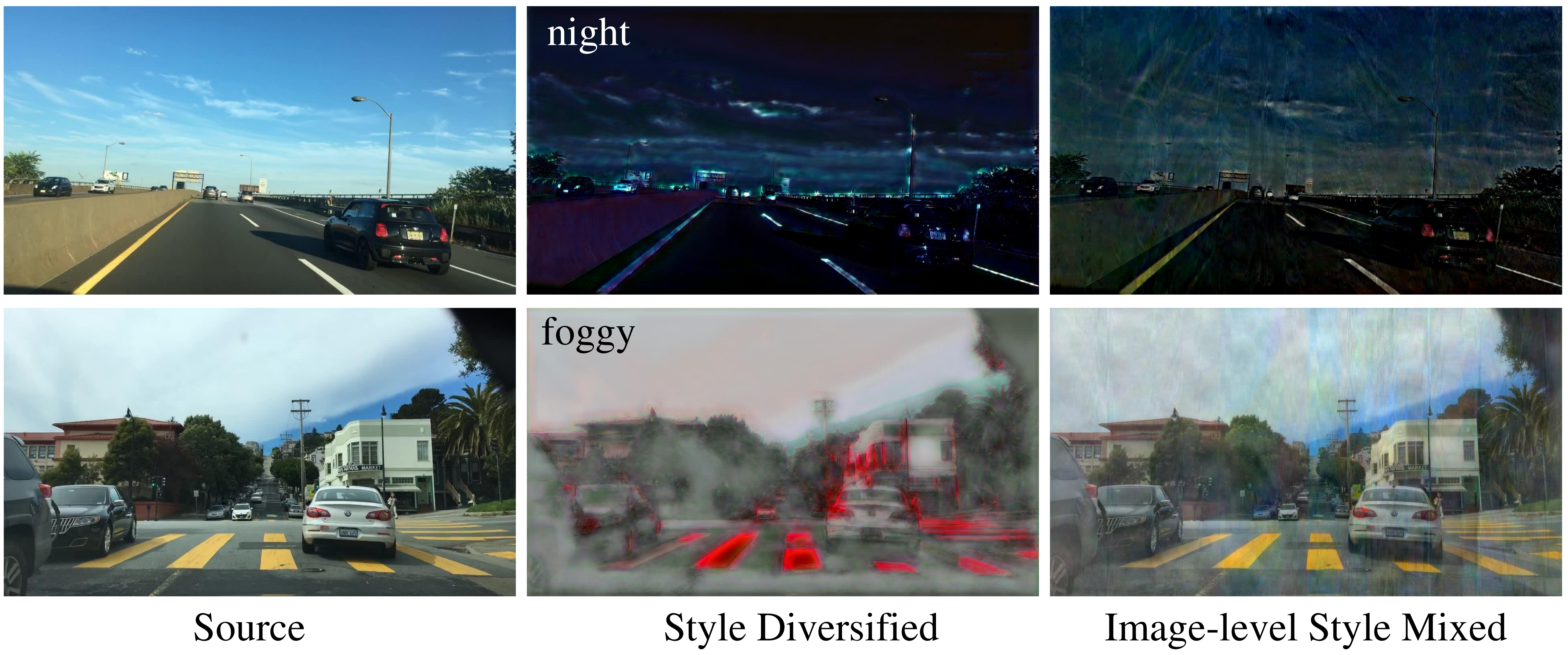}

   \caption{The visual results of the style generation and image-level style mixing, where the first and second columns correspond to the ${I}^{s}$ and the ${I}^{st}$, respectively. In the third column of ${I}^m$, while global style information has been mixed, the local styles, such as the object details and background textures, remain unchanged.}
   \label{fig:fft_compare}
\end{figure}

While the above style generation and image-level style mixing captures the style information of the diversified image ${I}^{st}$ to augment the source image ${I}^{s}$. It is still limited by the fact that the amplitude in the frequency domain only encodes global styles, \eg, brightness, and contrast. As shown in~\cref{fig:fft_compare}, although the ${I}^m$ visually appears to be heavily influenced by the style of the ${I}^{st}$,  the texture details between objects and the background, as well as the blending of multiple overlapping objects, remain in their original state. These factors weaken the ability of the detector to perceive full style information. To overcome this limitation, we consider that the statistical properties of the feature maps in the backbone network can extract local style information~\cite{adin, zhou2024mixstyle}. Therefore, we furthermore propose a feature-level style mixing augmentation scheme.

Before implementing this process, we note that existing VLM-based feature statistic transfer methods require the detector backbone to match the image encoder~\cite{fahes2023poda,vidit2023clip}, as they directly extract feature information from the backbone, requiring a shared network architecture. To avoid these architectural constraints, we employ a double-pipeline approach, as shown in~\cref{fig3}, where the ${I}^m$ and the ${I}^{st}$ are in parallel through the backbone. Building on this, we can proceed to design the feature-level style mixing based on statistics, without being restricted by the backbone.

\begin{figure}[t]
  \centering
   \includegraphics[width=1.0\linewidth]{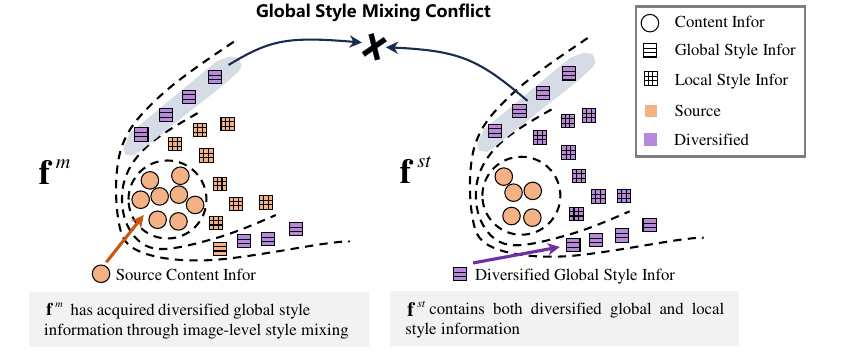}
   \caption{The semantic information space representation of $\mathbf{f}^m$ and $\mathbf{f}^{st}$ (source content information in $\mathbf{f}^{m}$ is complete, while part of it is missing in $\mathbf{f}^{st}$). Direct feature augmentation can cause global style mixing conflicts, which in turn disrupts the global style information in $\mathbf{f}^{m}$.}
   \label{fig:feature_mix}
\end{figure}

The feature statistical, \ie, means ${\mu \in \mathbb{R}^{C}}$ and standard deviations ${\sigma \in \mathbb{R}^{C}}$ are computed from the feature map $\mathbf{f} \in \mathbb{R}^{C \times H \times W}$. It is worth noting that source mixed image feature maps $\mathbf{f}^m$ have been augmented at the image level, meaning they are already mixed with global style information from ${I}^{st}$. However, since we use a double-pipeline process with ${I}^{st}$, the diversified image feature maps $\mathbf{f}^{st}$ also contain these global styles. As illustrated in \cref{fig:feature_mix}, direct statistical mixing could lead to excessive noise from redundant global styles, which has negative effects on this secondary style mixing. 

To allow smooth secondary style mixing, we construct a Gaussian Mixture Model (GMM) on the statistics $x^{st}=\{{\mu ^{st}},{\sigma ^{st}}\}$ of $\mathbf{f}^{st}$ to capture the most representative style information, focusing on the data within the highest-weight Gaussian distribution $p^ \star(x^{st})$. Formally, $p^ \star(x^{st})$ is define as follows.
\begin{equation}
\label{eq_gmm}
p^ \star(x^{st}) = {\cal N}(x^{st}|{\mu _{{k^ \star }}},{\Sigma _{{k^ \star }}}),\quad {k^ \star } = \arg \max {\pi _k},
\end{equation}
where $\pi _k$ denotes the mixing weight of the $k$-th Gaussian component, and we select 
the $k^ \star$-th Gaussian distribution that corresponds to the highest weight in $\pi _k$, with a mean of  ${\mu _{{k^ \star }}}$ and a covariance matrix ${\Sigma _{{k^ \star }}}$. The statistical style data ${x^{st\star}} \in p^ \star(x^{st})$ represents the most widely captured local style information in the ${I}^{st}$.

After obtaining ${x^{st\star}}$, with a portion of global style removed, performing feature-level augmentation can allow for a smooth secondary style mixing. This is achieved by mixing the feature statistics ${x^{st\star}}={\{\mu ^{st\star}},{\sigma ^{st\star}}\}$ and  ${x^m}={\{\mu ^{m}},{\sigma ^{m}}\}$ as follow:
\begin{equation}
\label{eq_mu_1}
{\mu ^{m'}} = (1 - \beta ){\mu ^m} + \beta {\mu ^{st\star}},\\
\end{equation}
\begin{equation}
\label{eq_sigma_1}
{\sigma ^{m'}} = (1 - \beta ){\sigma ^m} + \beta {\sigma ^{st\star}},
\end{equation}
where ${\mu ^{m'}}$ and ${\sigma ^{m'}}$ are the mixed statistics, and $\beta$ is sampled from the Beta distribution. In the following steps, we differ from MixStyle~\cite{zhou2024mixstyle}, which mixes the statistics of two images from the source domain and may disrupt object details. Since our smoothed style mixing reduces interference with object details, we further enhance semantic augmentation by introducing channel-wise randomly shuffling to the statistics, with ${\mu ^{m''} _{c}}  = {\mu ^{m'} _{\rm{shuffle} (c)}}$ and ${\sigma ^{m'} _{c}} = {\sigma ^{m'} _{\rm{shuffle} (c)}}$, where $c = 1,2, \ldots, C$. Finally, ${\mu ^{m''}}$ and ${\sigma ^{m''}}$ are used to restore the normalized feature $\mathbf{f}^m$ to obtain the dual style mixing feature $\mathbf{f}^{m\star}$. The smooth feature-level style mixing procedure is outlined in \cref{alg:r2p}.

\subsection{Generalization Training and Inference}
In the domain generalization training process, we first construct sufficient style prompts and fine-tune the StyleNet to produce diversified images from the source image. The generated diversified image with a randomly selected style is then mixed with the source image at the image level. Moreover, feature statistics from the double-pipeline detector backbone are randomly selected for smooth feature style mixing. Since only semantic augmentation is applied to the training data, the supervision signal is still based on the source image annotations with the original detection loss. In the inference phase, the detector, without any structural modifications, is evaluated on the unseen domain.

\SetCommentSty{mygreen}  %

\newcommand\mygreen{\textcolor{green}}
\begin{algorithm}[b]
    \caption{Smooth Feature-level Style Mixing}
    \label{alg:r2p}
    \KwIn{ Set $\mathbf{F}^m$ of image-level style mixed feature maps, set $\mathbf{F}^{st}$ of diversified image feature maps.}
    \KwOut{Dual style mixed feature maps $\mathbf{f}^{m\star}$.}
    \ForEach{$(\mathbf{f}^m \in \mathbf{F}^m$, $\mathbf{f}^{st} \in \mathbf{F}^{st})$}{
      ${\mu}^{m}, {\sigma}^{m} \gets \text{mean}(\mathbf{f}^m), \text{std}(\mathbf{f}^m)$\;
      ${\mu}^{st}, {\sigma}^{st} \gets \text{mean}(\mathbf{f}^{st}), \text{std}(\mathbf{f}^{st})$\;
      $p{(x^{st})} \gets \text{GMM}({\mu}^{st}, {\sigma}^{st}), \quad M=5$\;
      ${\mu}^{st\star}, {\sigma}^{st\star} \gets p^{\star}(x^{st}), \quad {k^{\star}} = \arg \max {\pi_k}$\;
      ${\mu}^{m{'}}, {\sigma}^{m{'}} \gets \text{mix}(\beta, {\mu}^{st\star}, {\sigma}^{st\star}, {\mu}^{m}, {\sigma}^{m})$\;
      ${\mu}^{m{''}}, {\sigma}^{m{''}} \gets \text{shuffle}({\mu}^{m{'}}, {\sigma}^{m{'}})$\;
      $\mathbf{f}^{m\star} \gets \text{norm}(\mathbf{f}^m) \times {\sigma}^{m{''}} + {\mu}^{m{''}}$\;
    }
\end{algorithm}

\section{Experiments}
\subsection{Datasets}

\noindent\textbf{Real to cartoon.} The PASCAL VOC dataset~\cite{everingham2010pascal} constructed from real-world scenes, serves as the source domain in this task, while the Clipart1k, Watercolor2k, and Comic2k datasets~\cite{inoue2018cross}, composed of cartoon images, are used as the unseen domains. A total of 16,551 images from VOC2007 and VOC2012 are set for the training dataset, and 5,000 images from VOC2007 are set for the source domain testing dataset. The cartoon datasets with Clipart1k, Watercolor2k, and Comic2k, include 1,000, 2,000, and 2,000 images, respectively, and serve as test sets. 
PASCAL VOC and Clipart1k share the same 20 classes, while Watercolor2k and Comic2k each contain 6 classes.

\noindent\textbf{Normal to adverse weather.} For the experiments on generalizing from normal weather to various adverse weather conditions, we use a dataset based on BDD100k~\cite{yu2020bdd100k}, which includes urban scene images with corresponding object detection annotations. Following previous research~\cite{vidit2023clip}, we use 19,395 daytime-sunny images (\textbf{DS}) for training and 8,313 for source domain testing. For testing on unseen domains, the dataset includes 3,501 dusk-rainy images (\textbf{DR}), 2,494 night-rainy images (\textbf{NR}), 26,158 night-clear images (\textbf{NC}), and 3,775 daytime-foggy (\textbf{DF}) images. All these subsets share the same 7 classes.

\subsection{Implementation Details}
We conduct experiments on both the one-stage, two-stage, and transformer-based detectors. For the one-stage detector, we select the popular YOLOv8-S~\cite{yolov8} with CSP-DarkNet backbone as the baseline, and following the original configuration. Set the image size to 640$\times$640, and use a batch size of 32 source images and 32 diversified images fed into the dual-pipeline. The initial learning rate is set to $10^{-2}$, and the model is trained for 100 epochs. For the two-stage detector, we use Faster R-CNN~\cite{ren2015faster} with ResNet-101 backbone as the baseline and maintain the original settings, a batch size of 8, and the initial learning rate of $10^{-2}$ for 18k iterations. For the transformer-based detector, we adopt RT-DTER~\cite{zhao2024detrs} with Swin Transformer-B backbone, adhering to the original configuration. We use the mean average precision (mAP) with an IoU threshold of 0.5 as the evaluation metric in all experiments. For the hyperparameter settings, the image-level style mixing parameters ${\gamma}_1$ and ${\gamma}_2$ are set for $0.5$ and $1.0$, the beta distribution parameter $\beta \sim Beta(0.1, 2.0)$ for feature-level style mixing, and $M = 5$ for the GMM. We conduct the experiments on 4 RTX3090 GPUs using PyTorch. Additional experimental setup details are provided in the supplementary materials.

\subsection{Results and Comparisons}

Besides the benchmark detector with source data, we compare the proposed LDDS against existing methods, including image augmentation methods like Div \cite{danish2024improving}, and feature augmentation methods based on VLMs, like CLIP-Gap~\cite{vidit2023clip} and {P\O DA}~\cite{fahes2023poda}. In cases where Faster R-CNN is used as the detector, we reference the reported performance from their original papers.

\begin{figure*}[ht]
\centering{\includegraphics[width=0.98\textwidth]{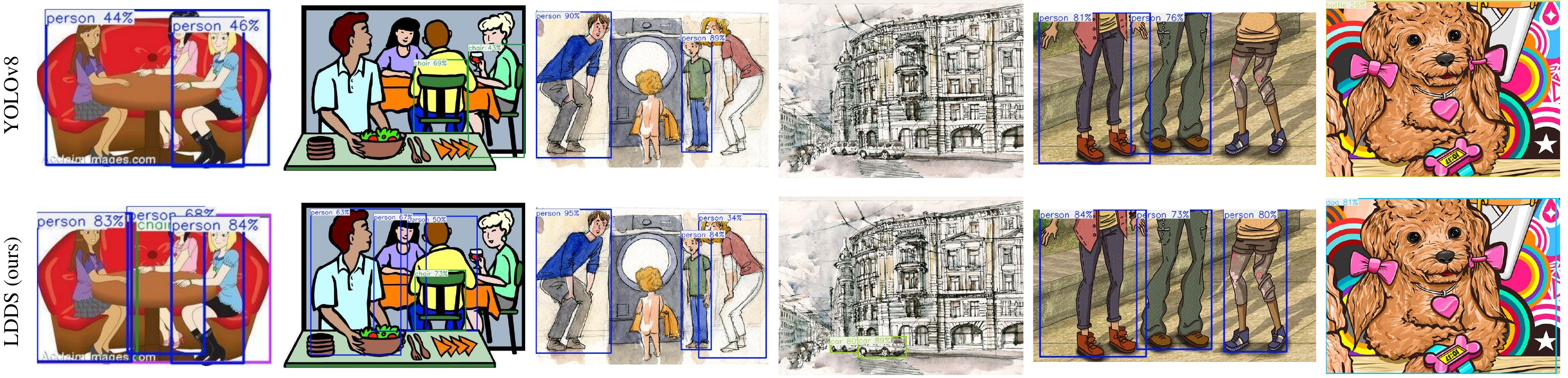}}
\caption{Qualitative results on the real to cartoon. From left to right, every two columns represent the samples of Clipart1k, Watercolor2k, and Comic2k, respectively. Our LDDS demonstrates better localization and classification of cartoon objects compared to the benchmark detector YOLOv8 \cite{yolov8}.}
\label{fig_cartoon}
\end{figure*}

\noindent\textbf{Real to cartoon.} To demonstrate the validity of LDDS, we first conduct experiments on real to cartoon scenes, characterized by a significant domain gap. We train on PASCAL VOC and test on Clipart1k, Watercolor2k, and Comic2k. As in~\cref{tab:cartoon_exp}, compared to the recent methods, NP~\cite{fan2023NP}, Div \cite{danish2024improving}, our LDDS achieves a gain of 14.4\%, 14.9\%, and 15.4\% on Faster R-CNN, with all performance metrics reaching optimal levels. In addition, in experiments with the one-stage detector YOLOv8, we outperform the source by 5.5\%, 9.2\%, and 14.1\%. The transformer-based detector RT-DETR exhibits more severe over-fitting on this task. LDDS preserves its performance in the source domain while achieving gains of 9.7\%, 12.5\%, and 14.6\% in the unseen domain. \cref{fig_cartoon} shows the detection results. It shows that even with the parameter-limited YOLOv8 model, our LDDS significantly enhances its ability to recognize and locate cartoon objects.

\begin{table}[h]
  \centering
  \fontsize{7.4pt}{8pt}\selectfont 
  \begin{tabular}{l|c|c|ccc}
    \toprule
    \noalign{\vspace{-1pt}}
    Method & Detector &VOC &Clipart &Watercolor &Comic \\
    \noalign{\vspace{-1pt}}
    \midrule
    FRCNN \cite{ren2015faster} & FRCNN &81.8 &25.7 &44.5 &18.9 \\
    NP \cite{fan2023NP} & FRCNN &79.2 &35.4 &53.3 &28.9 \\
    Div \cite{danish2024improving} & FRCNN & 80.1 &38.9 &57.4 &33.2 \\
    \rowcolor[gray]{0.93}
    LDDS (ours) & FRCNN & 81.5 &\bfseries40.1 &\bfseries59.4 &\bfseries34.3\\
    \hline
    \noalign{\vspace{+1pt}}
    YOLOv8 \cite{yolov8} & YOLO & 82.8 &34.2 &50.0 &19.8 \\
    \rowcolor[gray]{0.93}
    LDDS (ours) & YOLO & 82.5 &\bfseries39.7 &\bfseries59.2 &\bfseries33.9 \\
    \hline
    \noalign{\vspace{+1pt}}
    RT-DETR~\cite{zhao2024detrs}  & DETR & 85.6 &29.9 &48.7 &17.4 \\
    \rowcolor[gray]{0.93}
    LDDS (ours) & DETR & 85.9 &\bfseries38.6 &\bfseries61.2 &\bfseries32.0 \\
    \noalign{\vspace{-2pt}}
    \bottomrule
  \end{tabular}
  \caption{The results of SDG in real to cartoon (\%). ``FRCNN'' denotes Faster R-CNN, ``YOLO'' denotes YOLOv8, and ``DETR'' denotes RT-DETR. With VOC as the source domain and unseen domains being Clipart1k, Watercolor2k, and Comic2k, show that our LDDS achieves the best performance across all test datasets. This proves that dual style mixing is an effective way to enhance the generalization capability for cartoon data.}
  \label{tab:cartoon_exp}
\end{table}

\begin{table}[!h]
  \centering
  \renewcommand{\arraystretch}{1.0}
  \fontsize{7.5pt}{8pt}\selectfont
  \begin{tabular}{l|c|c|cccc}
    \toprule
    \noalign{\vspace{-1pt}}
    Method  & Detector &DS &NC &DR &NR &DF\\
    \noalign{\vspace{-1pt}}
    \midrule
    FRCNN~\cite{ren2015faster} & FRCNN &51.8 &38.9 &30.0 &15.7 &33.1 \\
    IterNorm~\cite{IterNorm} & FRCNN &43.9 &29.6 &22.8 &12.6 &28.4\\
    ISW~\cite{ISW} & FRCNN &51.3 &33.2 &25.9 &14.1 &31.8\\
    S-DGOD~\cite{wu2022single} & FRCNN &56.1 &36.6 &28.2 &16.6 &33.5\\
    CLIP-Gap~\cite{vidit2023clip} & FRCNN &51.3 &36.9 &32.3 &18.7 &38.5\\
    {P\O DA}~\cite{fahes2023poda} & FRCNN & - &43.4 &40.2 &20.5 &\bfseries44.4\\
    Div~\cite{danish2024improving} & FRCNN &52.8 &42.5 &38.1 &24.1 &37.2\\ %
    \rowcolor[gray]{0.93}
    LDDS (ours)  & FRCNN &52.6 &\bfseries44.1 &\bfseries43.2 &\bfseries27.8 &39.3\\
    \hline
    \noalign{\vspace{+1pt}}
    YOLOv8~\cite{yolov8} & YOLO &57.3 &43.1 &27.9 &13.7 &35.7\\
    \rowcolor[gray]{0.93}
    LDDS (ours) & YOLO &57.5 &\bfseries45.6 &\bfseries32.2 &\bfseries19.1 &\bfseries37.7\\
    \hline
    \noalign{\vspace{+1pt}}
    RT-DETR~\cite{zhao2024detrs}  & DETR &56.9 &40.8 &35.2 &21.6 &41.3\\
    \rowcolor[gray]{0.93}
    LDDS (ours)  & DETR &57.0 &\bfseries46.3 &\bfseries41.9 &\bfseries27.4 &\bfseries45.1\\
    \noalign{\vspace{-2pt}}
    \bottomrule
  \end{tabular}
  \caption{Results of normal to adverse weather on the BDD100k dataset (\%).  For the YOLOv8 and RT-DETR, we achieve consistent performance improvements and make significant progress in the most challenging scenario of night-rainy. For the Faster R-CNN,  LDDS outperforms all methods on three datasets. While {P\O DA}~\cite{fahes2023poda} achieves superior results on the daytime-foggy, its usage is restricted by backbone network limitations.}
  \label{tab:bdd100k_compare}
\end{table}

\begin{figure}[h]
  \centering
   \includegraphics[width=1.0\linewidth]{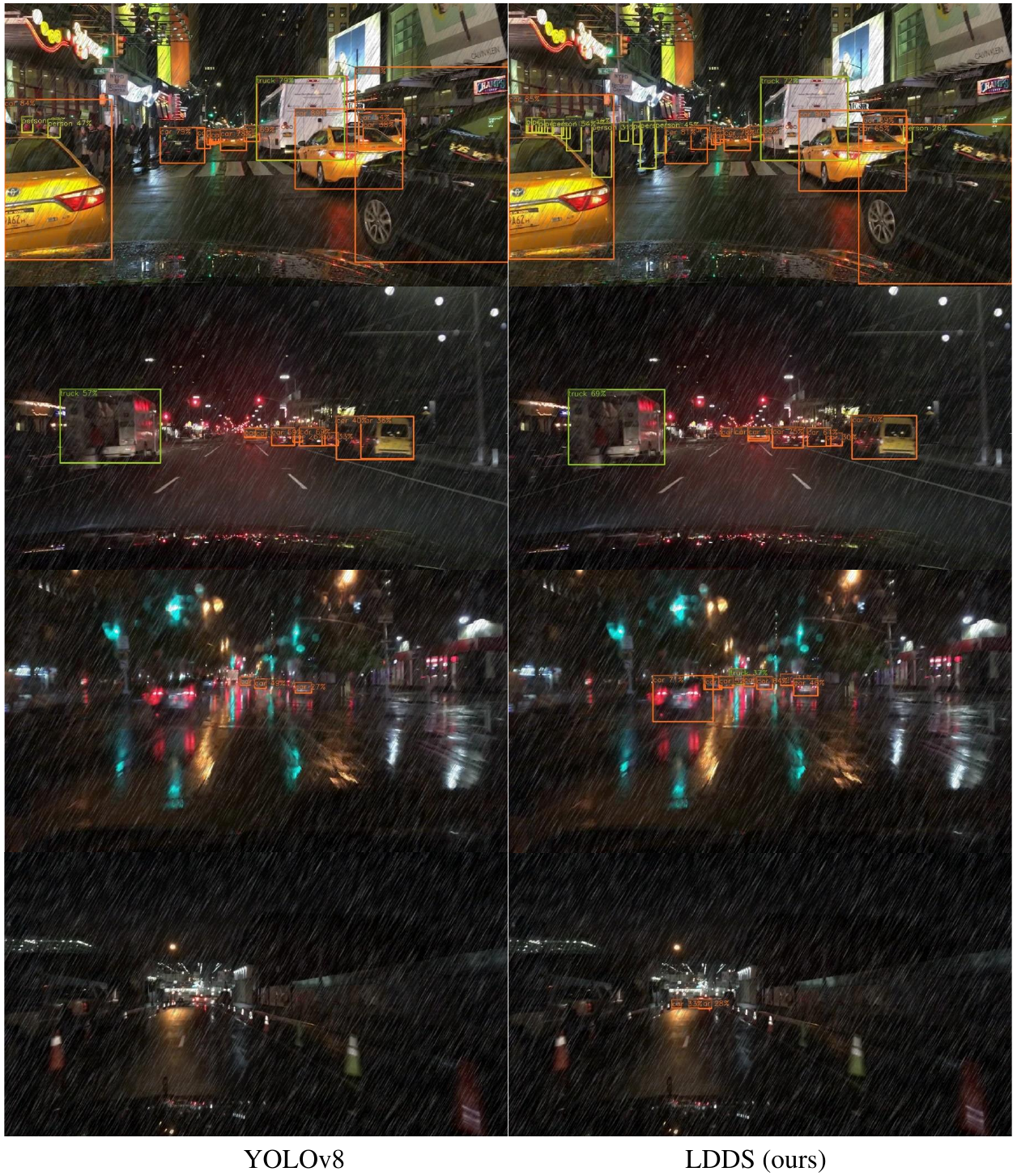}

   \caption{Qualitative results on the night-rainy dataset. We compare the detection results of YOLOv8~\cite{yolov8} and LDDS on the most challenging night-rainy dataset, showing that our method accurately detects the objects in this scenario.}
   \label{fig:compare_night_yolo}
\end{figure}

\noindent\textbf{Normal to adverse weather.} The results presented in~\cref{{tab:bdd100k_compare}} clearly indicate that LDDS is among the top-performing ones. For the Faster R-CNN detector, compared to the state-of-the-art method Div \cite{danish2024improving}, we achieve a gain of 1.6\%, 5.1\%, 3.7\%, and 2.1\% on the night-clear, dusk-rainy, night-rainy, and daytime-foggy datasets, respectively. These performance boosts are largely due to LDDS effectively utilizing the rich weather domain information available in VLM for image and feature style mixing. For the YOLOv8 detector, our LDDS also achieves performance gains, particularly in the night-rainy dataset, where we see a 5.4\% increase. For the RT-DETR detector, LDDS achieves significant gains across all unseen domains, even with a higher baseline. \cref{fig:compare_night_yolo} illustrates the detection results in night-rainy, revealing that objects affected by rain streaks and inadequate lighting are still accurately detected.  This highlights the effectiveness of implementing language-driven style information in the image and style augmentation strategy for this field.

\subsection{Ablation Study}
To elucidate the contribution of each component in our method, we conduct a series of experiments for an ablation study. We select the most representative task of normal to adverse weather and use the one-stage object detector YOLOv8 as the benchmark.

\begin{figure}[t]
  \centering
   \includegraphics[width=1.0\linewidth]{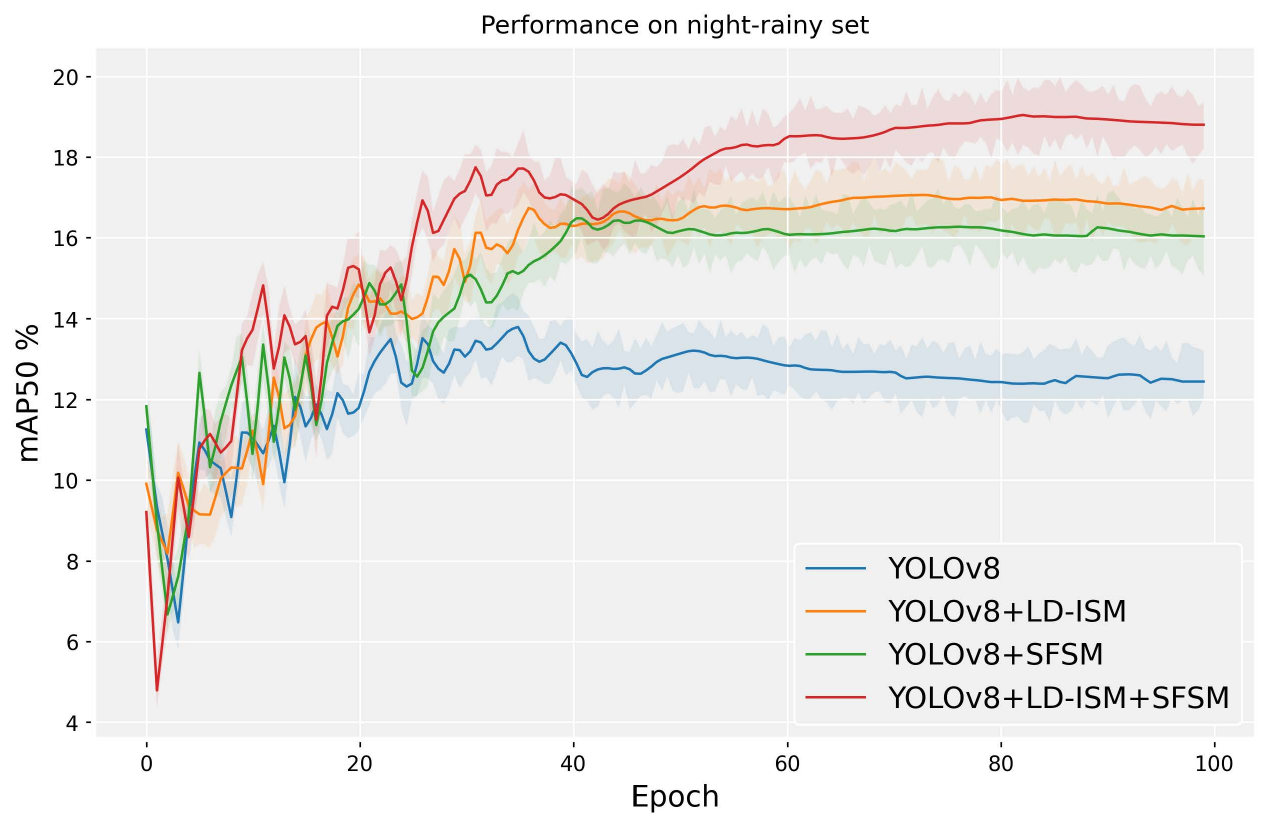}

   \caption{Results for model performance on the night-rainy dataset. We perform experiments on the benchmark detector, adding each proposed component and illustrating the boundary error. As shown in the figure, each component exhibits a positive impact on the overall performance trend.}
   \label{fig:ablation_night_yolo}
\end{figure}

\noindent\textbf{Quantitative ablation.} Since both image-level and feature-level style mixing in LDDS utilize diversified images from style generation, in the ablation setting, \textbf{LDDS-w/o LD-ISM} denotes the removal of the image style mixing pipeline, which includes language-driven diversified image inputs and subsequent image-level style mixing. \textbf{LDDS-w/o SFSM} denotes the removal of the smooth feature style mixing pipeline, which involves diversified images feature extraction and smooth feature-level style mixing. \textbf{LDDS-w/o p-GMM} denotes the removal of the smoothing process for feature-level style mixing. \textbf{LDDS (full)} denotes the full model. As observed in~\cref{tab:ablation}, the results demonstrate that each component we design contributes to the overall framework, as evidenced by varying degrees of performance degradation upon their removal. We also provide ablation results on RT-DETR to better support this view. \cref{fig:ablation_night_yolo} depicts the performance curves by incorporating components into the baseline detector during training.

\renewcommand{\arraystretch}{0.9}
\begin{table}[t]
  \centering
  \renewcommand{\arraystretch}{1.0} %
  \fontsize{7.2pt}{8pt}\selectfont
  \begin{tabular}{l|c|c|cccc}
    \toprule
    \noalign{\vspace{-1pt}}
    Method & Detector &DS &NC &DR &NR &DF\\
    \noalign{\vspace{-1pt}}
    \midrule
    YOLOv8~\cite{yolov8}  & YOLO &57.3 &43.1 &27.9 &13.7 &35.7 \\
    LDDS-w/o LD-ISM  & YOLO &56.3 &43.4 &28.7 &16.3 &36.4 \\
    LDDS-w/o SFSM  & YOLO &58.2 &44.8 &29.1 &16.9 &36.5\\
    LDDS-w/o p-GMM  & YOLO &57.4 &45.4 &31.0 &18.0 &\bfseries37.9\\
    \rowcolor[gray]{0.93}
    LDDS (full)    & YOLO &57.5 &\bfseries45.6 &\bfseries32.2 &\bfseries19.1 &37.7\\
    \hline
    \noalign{\vspace{+1pt}}
    RT-DETR~\cite{zhao2024detrs}  & DETR &56.9 &40.8 &35.2 &21.6 &41.3\\
    LDDS-w/o LD-ISM & DETR &56.1 &42.3 &38.2 &23.4 &43.0\\
    LDDS-w/o SFSM  & DETR &57.2 &44.8 &39.6 &25.6&43.6\\
    LDDS-w/o p-GMM  & DETR &57.4 &45.1 &41.5 &26.6 &44.8\\
    \rowcolor[gray]{0.93}
    LDDS (full)  & DETR &57.0 &\bfseries46.3 &\bfseries41.9 &\bfseries27.4 &\bfseries45.1\\
    \noalign{\vspace{-2pt}}
    \bottomrule
  \end{tabular}
  \caption{Ablation studies on the model components (\%).}
  \label{tab:ablation}
\end{table}

\noindent\textbf{Analysis of image augmentation methods.} We analyze the impact of various frequency-domain image-level augmentation methods on the proposed LDDS. \cref{tab:ablation_fft} presents the experimental results for the normal to adverse weather task. We compare three widely used methods including Fourier Corruption~\cite{cugu2022attention}, Fourier Transfer~\cite{yang2020fda}, and Fourier Mix~\cite{xu2021fourier}. The results clearly show that our language-driven image-level style mixing achieves the best performance improvement. This is attributed to LD-ISM capturing rich global style information for image augmentation while minimizing the destruction of object details.

\renewcommand{\arraystretch}{0.9}
\begin{table}[!t]
  \centering
  \renewcommand{\arraystretch}{1.0} %
  \footnotesize 
  \begin{tabular}{l|c|cccc}
    \toprule
    \noalign{\vspace{-1pt}}
    Component &DS &NC &DR &NR &DF\\
    \noalign{\vspace{-1pt}}
    \midrule
    LDDS-w/o LD-ISM &56.3 &43.4 &28.7 &16.3 &36.4 \\
    + Fourier Corruption~\cite{cugu2022attention} &57.8 &42.1 &30.1 &16.7 &35.1 \\
    + Fourier Transfer~\cite{yang2020fda} &54.3 &43.8 &29.6 &17.3 &36.6\\
    + Fourier Mix~\cite{xu2021fourier} &56.0 &44.9 &29.5 &17.8 &35.6\\
    \rowcolor[gray]{0.93}
    + LD-ISM (ours)   &57.5 &\bfseries45.6 &\bfseries32.2 &\bfseries19.1 &\bfseries37.7\\
    \noalign{\vspace{-2pt}}
    \bottomrule
  \end{tabular}
  \caption{Results comparison of different image augmentation methods (\%). We reported the performance impact of implementing various frequency augmentation components within the proposed framework.}
  \label{tab:ablation_fft}
\end{table}

\noindent\textbf{Analysis of feature augmentation methods.} \cref{{tab:ablation_adin}} shows the impact of other feature augmentation methods on performance compared to SFSM within our LDDS. Both ADIN~\cite{adin} and MixStyle~\cite{zhou2024mixstyle}, which also utilize statistical information from feature maps, effectively enhance generalization performance within source domain data. However, under our framework, SFSM shows superior adaptability, benefiting from the optimization of pre-mixed features $\mathbf{f}^m$. This proves that SFSM reduces conflicts arising from global style mixing, resulting in a smoother dual style mixing that achieves semantic augmentation.

\renewcommand{\arraystretch}{0.9}
\begin{table}[t]
  \centering
  \renewcommand{\arraystretch}{1.0} %
  \footnotesize 
  \begin{tabular}{l|c|cccc}
    \toprule
    \noalign{\vspace{-1pt}}
    Component &DS &NC &DR &NR &DF\\
    \noalign{\vspace{-1pt}}
    \midrule
    LDDS-w/o SFSM   &58.2 &44.8 &29.1 &16.9 &36.5\\
    + ADIN~\cite{adin} &57.0 &44.6 &30.1 &17.4 &33.1 \\
    + MixStyle~\cite{zhou2024mixstyle} &57.6 &\bfseries45.8 &29.8 &18.0 &35.9\\
    \rowcolor[gray]{0.93}
    + SFSM (ours)   &57.5 &45.6 &\bfseries32.2 &\bfseries19.1 &\bfseries37.7\\
    \noalign{\vspace{-2pt}}
    \bottomrule
  \end{tabular}
  \caption{Results comparison of different feature augmentation methods (\%). We reported the results of applying commonly used feature statistic style methods, ADIN~\cite{adin} and MixStyle~\cite{zhou2024mixstyle}, within the proposed framework.}
  \label{tab:ablation_adin}
\end{table}

\begin{figure}[h]
  \centering
   \includegraphics[width=0.8\linewidth]{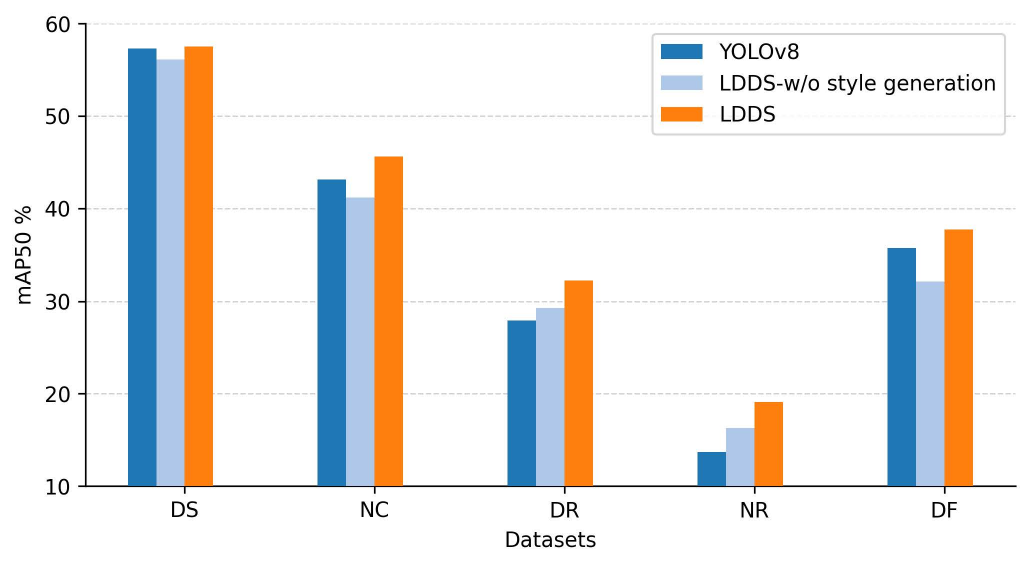}
   \caption{The reported effect of style generation on the performance of each test set for normal to adverse weather tasks.}
   \label{fig:compare_CLIPstyle}
\end{figure}

\noindent\textbf{Effect of style generation.} Utilizing style generation to introduce diverse style information is one of the key motivations of our LDDS. To further evaluate its significance, we perform an analysis by decoupling its components. We remove the branch handling the diversified images and replace it with the source domain images. As shown in~\cref{fig:compare_CLIPstyle}, both image-level and feature-level mixing lead to improved generalization performance in adverse weather scenarios, such as dusk-rainy and night-rainy. However, the lack of more extensive information limited its effectiveness in other conditions, like the daytime-foggy. This underscores the importance of the language-driven approach in ensuring both specificity and broad generalization.
\section{Conclusion}
In this work, we propose a simple SDG object detection method called LDDS, which allows for diversifying the source domain based on VLM. We obtain explicit unseen domain style information from VLM using style prompts. We propose dual style mixing to augment both images and features by the style information, and utilizing a smooth strategy to avoid conflicts from double rounds of style mixing. LDDS addresses key challenges in SDG object detection, utilizing the semantic information from VLMs without being constrained by the detector backbone architecture. Extensive experiments show that LDDS surpasses existing image and feature augmentation methods, offering a novel research paradigm for this field.

{
    \small
    \bibliographystyle{ieeenat_fullname}
    \bibliography{main}
}

\clearpage
\setcounter{page}{1}

In the supplementary material, \cref{sec:Parameter} begins with a sensitivity analysis of the two mixing parameters in dual-style mixing. \cref{sec:Experimental} provides additional implementation details along with class-level experimental results. \cref{sec:Visualization} presents the visualization results, which include a comparative of image augmentation methods and qualitative comparison based on different detector baselines. Finally, \cref{sec:Discussion} discusses the importance of detector backbone networks and the limitations identified in the current work.

\section{Parameter Sensitivity Analysis}
\label{sec:Parameter}
As shown in \cref{tab:para_mix}, we analyze the sensitivity of the style mixing parameters in the proposed LDDS and provide quantitative results on the daytime-sunny to night-rainy task. For the image-level style mixing parameter $\omega \sim U({\gamma}_1,{\gamma}_2)$ and feature-level style mixing parameter $\beta \sim Beta(\beta _1, \beta _2)$, we first fixed $\beta _{1,2}$ and then tried experimented ($\rm{1}^{st}$ to $\rm{7}^{th}$ lines) with adjusting $\omega$ within different ranges. We observe that a higher mixing ratio leads to improved performance gains, thus due to the LD-ISM mitigating object detail degradation of diversified images. However, for the sake of preserving diverse styles to guarantee generalization, a well-chosen higher mixing ratio is preferable (${\gamma}_1 = 0.5$, ${\gamma}_2 = 1.0$).

\renewcommand{\arraystretch}{0.9}
\begin{table}[h]
  \centering
  \renewcommand{\arraystretch}{1.0} %
  \footnotesize 
  \begin{tabular}{p{0.6cm}c|p{0.6cm} c|cc}
    \toprule
    \noalign{\vspace{-1pt}}
    \centering${\gamma}_1$ & ${\gamma}_2$ & \centering$\beta _1$ &$\beta _2$ & $\rm{mAP}_{0.5}$ &$\rm{mAP}_{0.5:0.95}$\\
    \noalign{\vspace{-1pt}}
    \midrule
    \centering0.0 &0.5 & \multicolumn{2}{c|}{\multirow{7}{*}{\parbox{1.5cm}{\centering $\beta_1 = 0.1$, \\ $\beta_2 = 1.0$}}}   &16.1 &9.0 \\
    \centering0.0 &0.8 & & &16.3 &9.3 \\
    \centering0.0  &1.0 & & &16.9 &9.6\\
    \centering0.3 &1.0 & & &17.6 &9.9\\
    \centering0.5 &1.0 & & &\bfseries18.4 &\bfseries10.4\\
    \centering0.8 &1.0 & & &16.9 &9.7\\
    \centering1.0 &1.0 & & &15.9 &8.9\\
    \hline
    \noalign{\vspace{1pt}}
    \multicolumn{2}{c|}{\multirow{8}{*}{\parbox{1.5cm}{\centering ${\gamma}_1 = 0.5$, \\ ${\gamma}_2 = 1.0$}}}  &\centering0.1 &0.1 &18.3 &10.4 \\
     & &\centering0.1 &1.0 &18.4 &10.4 \\
     & &\centering0.1 &2.0 &\bfseries19.1 &\bfseries11.2\\
     & &\centering1.0 &1.0 &18.3 &10.3\\
     & &\centering0.8 &1.2 &16.8 &9.3\\
     & &\centering2.0 &0.5 &16.4 &9.2\\
     & &\centering0.5 &2.0 &18.7 &10.8\\
     & &\centering2.0 &2.0 &17.0 &9.4\\
    \noalign{\vspace{-2pt}}
    \bottomrule
  \end{tabular}
  \caption{Quantitative results on the daytime-sunny to the night-rainy dataset of different style mixing parameters settings of $\gamma _1$, $\gamma _2$, $\beta _1$, and $\beta _2$ (\%). }
  \label{tab:para_mix}
\end{table}

For the feature-level style mixing parameter $\beta$, we investigate the detection performance across different mixing parameters, aiming to find a suitable balance of feature augmentation with the mixed image-level style information. Specifically, we compare Beta distributions with extreme bias (\eg, $\beta _1 = 0.1$, $\beta _2 = 0.1$), asymmetrical bias ($\beta _1 = 0.1$, $\beta _2 = 1.0$), moderate bias ($\beta _1 = 0.8$, $\beta _2 = 1.2$), uniform mixing ($\beta _1 = 2.0$, $\beta _2 = 2.0$), and edge cases ($\beta _1 = 2.0$, $\beta _2 = 0.5$). In \cref{tab:para_mix}, we observe that applying a proper bias towards the source mixed image feature yielded the best experimental results ($\beta _1 = 0.1$, $\beta _2 = 2.0$), as it aids in preserving mixed image-level style information and making the feature-level style mixing a well-positioned secondary augmentation step. 

\begin{figure*}[ht]
\centering{\includegraphics[width=0.98\textwidth]{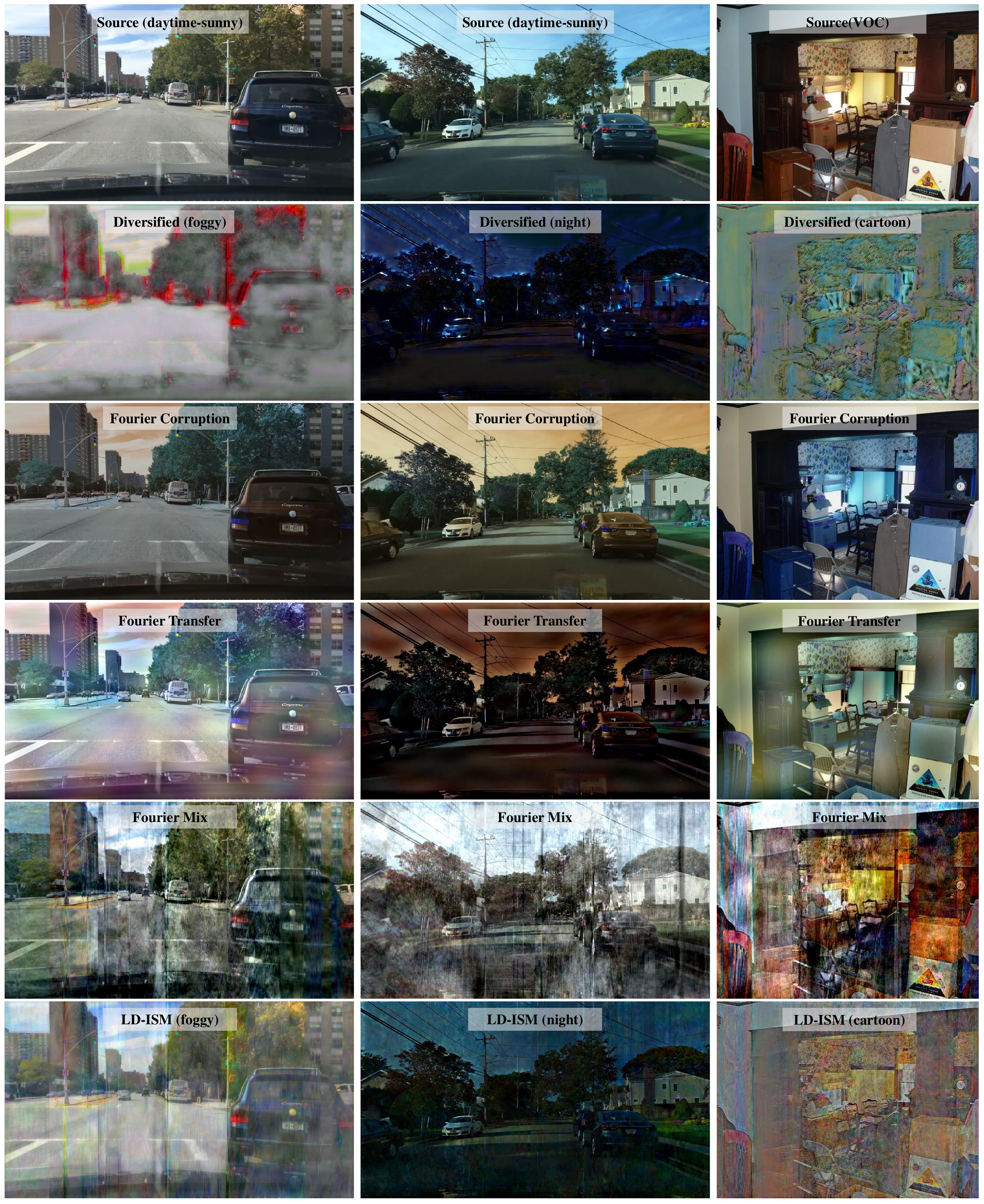}}
\caption{Illustration of the image augmentation visualization results, including Diversified \cite{kwon2022clipstyler}, Fourier Corruption \cite{cugu2022attention}, Fourier Transfer \cite{yang2020fda}, Fourier Mix \cite{xu2021fourier}, and LD-ISM (ours). }
\label{fig:fft}
\end{figure*}

\section{Additional Experimental Details}
\label{sec:Experimental}
\subsection{Implementation Details}
In the experiments based on one-stage, two-stage, and transformer-based detectors, the style generation component adhered to a consistent setup. Our style generation module, composed of a ResNet-101 CLIP image encoder, a CLIP text encoder, and a StyleNet built on UNet. To generate diversified images, we craft style prompts to suit different SDG object detection tasks. The source domain is assign the prompt \texttt{a photo}, whereas for the normal to adverse task, the style prompt was set as \texttt{a photo of the road in the \{weather\}}, where \texttt{weather} was substituted with \texttt{rainy}, \texttt{night}, or \texttt{foggy}, with appropriate grammatical modifications. For the real to cartoon task, the style prompt is set as \texttt{a photo in a \{cartoon\} style}, where the \texttt{cartoon} is filled with \texttt{clipart}, \texttt{comic}, or \texttt{watercolor}.

For feature-level style mixing, we select early layers in the backbone network whose feature statistics encode image style information. Due to architectural variations, we adapt layer selection configurations separately for each detector. In YOLOv8-S, we extract features from the output of the 4th convolutional module of CSP-DarkNet, while in Faster R-CNN, we extract features from the output of the 3rd convolutional module of ResNet-101. In RT-DETR, we implement the feature-level style mixing on the tokens from the output of the 3rd Swin Transformer block. To clarify, in RT-DETR, we substitute the original CNN-based HGNetv2 with Swin Transformer-B to demonstrate that our proposed method is equally applicable to transformer-based backbones.

\subsection{Class-wise Experimental Results}
We present details of the class-wise ablation studies results for daytime-sunny to night-rainy~(\cref{tab:night}), dusk-rainy~(\cref{tab:dusk}), and daytime-foggy~(\cref{tab:foggy}), offering in-depth insights into each class. As shown in~\cref{tab:night}, in the challenging night-rainy dataset, LDDS achieves gains across all categories. Particularly in common categories, \eg, person and car, which is particularly meaningful for real-world applications. 

\renewcommand{\arraystretch}{0.9}
\begin{table}[h]
  \centering
  \renewcommand{\arraystretch}{1.0} %
  \footnotesize 
  \setlength\tabcolsep{2pt}
  \begin{tabular}{@{}l|cccccccc}
    \toprule
    \noalign{\vspace{-1pt}}
    Method &Bus &Bike &Car &Motor &Person &Rider &Truck &mAP\\
    \noalign{\vspace{-1pt}}
    \midrule
    YOLOv8~\cite{yolov8} &24.7 &3.9 &33.4 &0.3 &8.7 &3.9 &21.3 &13.7\\
    LDDS-w/o LD-ISM &30.5 &4.5 &38.1 &0.6 &10.5 &5.1 &24.6 &16.3\\
    LDDS-w/o SFSM &31.1 &6.0 &40.4 &0.5 &9.7 &5.8 &25.3 &16.9\\
    LDDS-w/o p-GMM &32.2 &5.7 &40.1 &3.5 &10.1 &7.3 &27.0 &18.0\\
    \rowcolor[gray]{0.93}
    LDDS (full)   &\bfseries33.5 &\bfseries6.1 &\bfseries42.0 &\bfseries3.6 &\bfseries11.9 &\bfseries8.3 &\bfseries28.2 &\bfseries19.1\\
    \noalign{\vspace{-2pt}}
    \bottomrule
  \end{tabular}
  \caption{Ablation studies on the daytime-sunny to night-rainy (\%).}
  \label{tab:night}
\end{table}

\renewcommand{\arraystretch}{0.9}
\begin{table}[h]
  \centering
  \renewcommand{\arraystretch}{1.0} %
  \footnotesize 
  \setlength\tabcolsep{2pt}
  \begin{tabular}{@{}l|cccccccc}
    \toprule
    \noalign{\vspace{-1pt}}
    Method &Bus &Bike &Car &Motor &Person &Rider &Truck &mAP\\
    \noalign{\vspace{-1pt}}
    \midrule
    YOLOv8~\cite{yolov8} &33.5 &10.6 &67.5 &5.2 &22.8 &12.9 &42.4 &27.9\\
    LDDS-w/o LD-ISM  &37.0 &9.7 &68.1 &6.7 &22.8 &11.3 &44.9 &28.7\\
    LDDS-w/o SFSM &35.8 &9.4 &68.3 &\bfseries9.5 &23.7 &12.1 &44.8 &29.1\\
    LDDS-w/o p-GMM &41.2 &12.2 &69.3 &9.1 &27.3 &11.3 &\bfseries46.9 &31.0\\
    \rowcolor[gray]{0.93}
    LDDS (full)  &\bfseries41.8 &\bfseries15.2 &\bfseries70.3 &9.4 &\bfseries29.1 &\bfseries12.5 &46.8 &\bfseries32.2\\
    \noalign{\vspace{-2pt}}
    \bottomrule
  \end{tabular}
  \caption{Ablation studies on the daytime-sunny to dusk-rainy (\%).}
  \label{tab:dusk}
\end{table}

As indicated in~\cref{tab:dusk}, the class-wise performance gains in the dusk-rainy dataset are also steady. It is noteworthy in~\cref{tab:foggy}, the performance on the daytime-foggy dataset showed fluctuations. This is primarily caused by the dense fog in the image, which introduces greater interference with the objects. Setting specific style mixing parameters is effective for performance in foggy scenarios. However, for ease of deployment, our method uses identical parameters across all test datasets, without specific tuning for individual cases.

\renewcommand{\arraystretch}{0.9}
\begin{table}[h]
  \centering
  \renewcommand{\arraystretch}{1.0} %
  \footnotesize 
  \setlength\tabcolsep{2pt}
  \begin{tabular}{@{}l|cccccccc}
    \toprule
    \noalign{\vspace{-1pt}}
    Method &Bus &Bike &Car &Motor &Person &Rider &Truck &mAP\\
    \noalign{\vspace{-1pt}}
    \midrule
    YOLOv8~\cite{yolov8} &36.9 &26.6 &63.5 &25.1 &37.3 &37.0 &23.4 &35.7\\
    LDDS-w/o LD-ISM &35.0 &28.4 &61.5 &\bfseries28.3 &38.1 &40.4 &22.6 &36.4\\
    LDDS-w/o SFSM &36.4 &28.7 &63.2 &28.1 &37.1 &37.5 &24.6 &36.5\\
    LDDS-w/o p-GMM &\bfseries36.4 &28.7 &65.1 &28.1 &40.1 &38.5 &\bfseries28.3 &\bfseries37.9\\
    \rowcolor[gray]{0.93}
    LDDS (full)   &36.2 &\bfseries28.9 &\bfseries65.2 &27.4 &\bfseries40.9 &\bfseries41.6 &24.8 &37.7\\
    \noalign{\vspace{-2pt}}
    \bottomrule
  \end{tabular}
  \caption{Ablation studies on the daytime-sunny to daytime-foggy (\%).}
  \label{tab:foggy}
\end{table}

\section{Additional Visualization Results}
\label{sec:Visualization}
\subsection{Image augmentation visualization results}
We analyze the effects of image augmentation on SDG object detection by providing visual comparisons of different augmentation methods on source images, including Fourier Corruption~\cite{cugu2022attention}, Fourier Transfer~\cite{yang2020fda}, and Fourier Mix~\cite{xu2021fourier}. In addition, we provided the augmentation results from LD-ISM, as well as diversified images from the style generation module driven by corresponding style prompts. As shown in~\cref{fig:fft}, for the first column, we noted that visually diversified images often carried significant foggy information but led to object detail blurring. Fourier Corruption results in only minor visual alterations and Fourier Transfer creates effects resembling colored light shadows. Fourier Mix generated noise rather than image augmentation, while some noise of the image might yield robustness, it also risks concealing negative optimization. LD-ISM achieved more targeted augmentation, effectively preserving object details during foggy style information transfer.  For the other two columns, Fourier Transfer produced notable style changes in nighttime scenes, suggesting an avenue for further investigation. Nevertheless, achieving such enhancements relies on diversified image inputs. For cartoon images, LD-ISM also successfully augmented the visual quality while maintaining semantic integrity.

\begin{figure}[!t]
  \centering
   \includegraphics[width=1.0\linewidth]{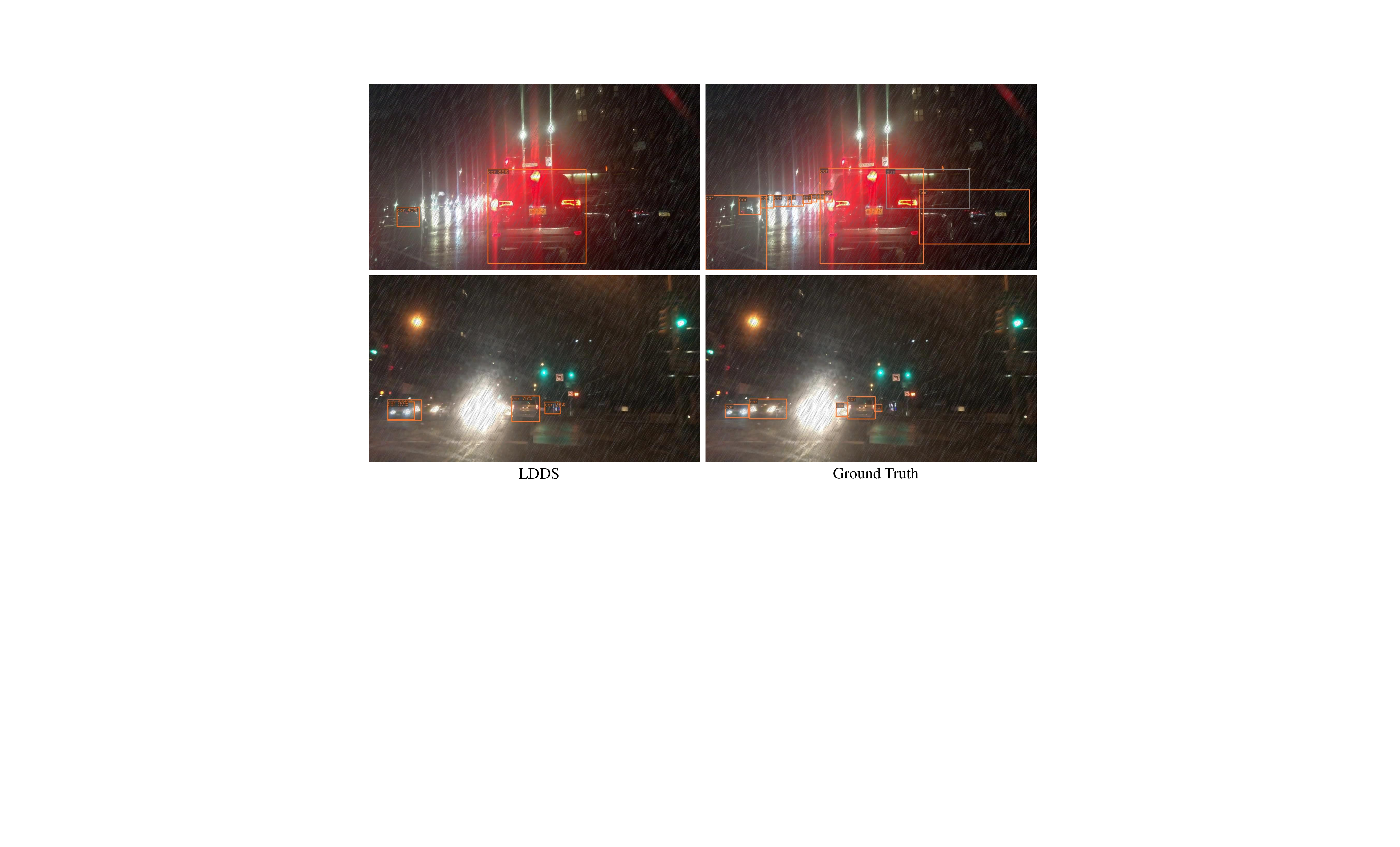}

   \caption{Visualization results on the target domain of daytime-sunny to night-rainy for LDDS (ours), and Ground Truth. Direct high-beam headlights create considerable obstacles for detection tasks.}
   \label{fig:night_limit}
\end{figure}

\subsection{Qualitative comparison}
\begin{figure*}[ht]
\centering{\includegraphics[width=0.98\textwidth]{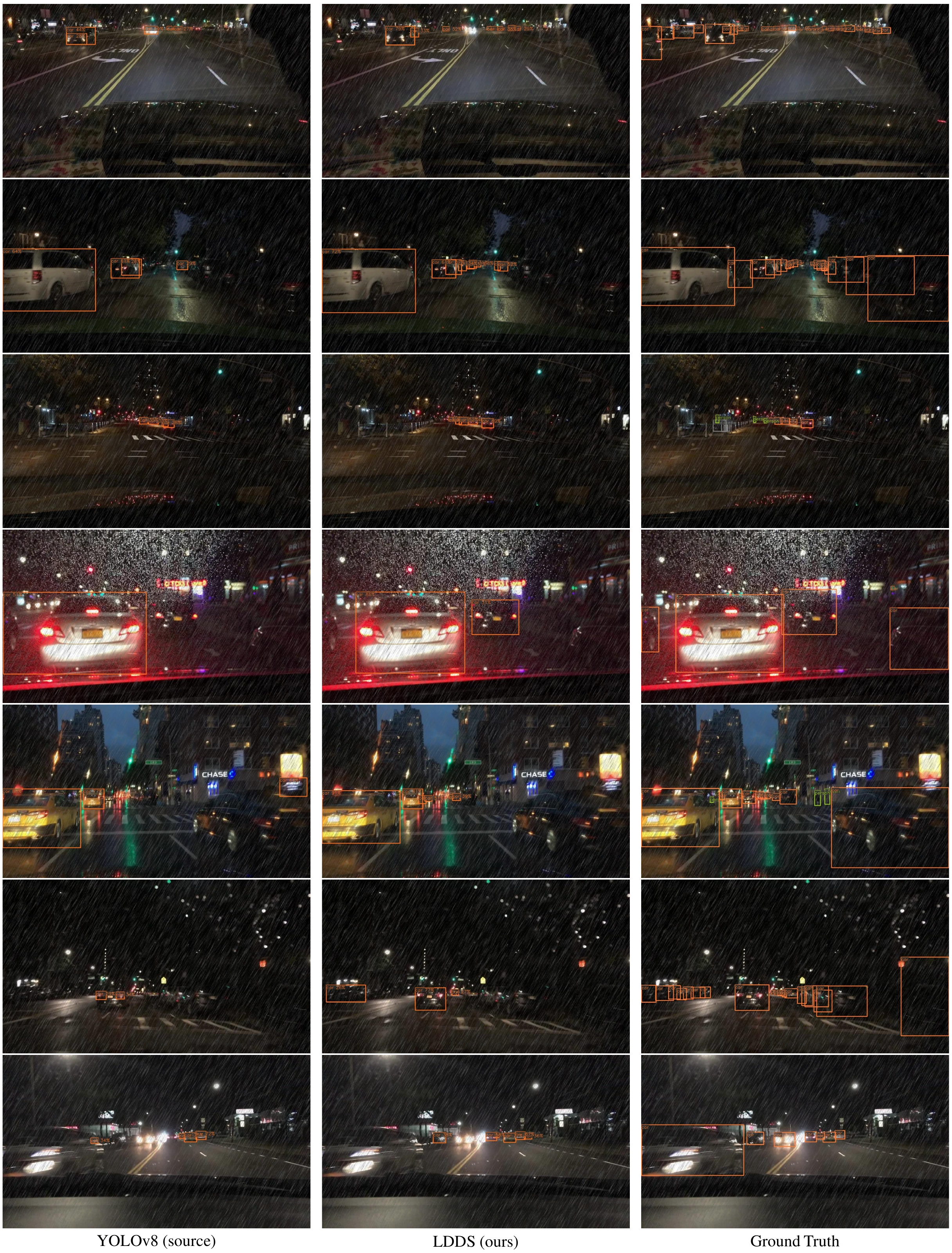}}
\caption{Additional qualitative results on the target domain of daytime-sunny to night-rainy for YOLOv8 (source) \cite{yolov8}, LDDS (ours), and Ground Truth.}
\label{fig:night}
\end{figure*}

\begin{figure*}[ht]
\centering{\includegraphics[width=0.98\textwidth]{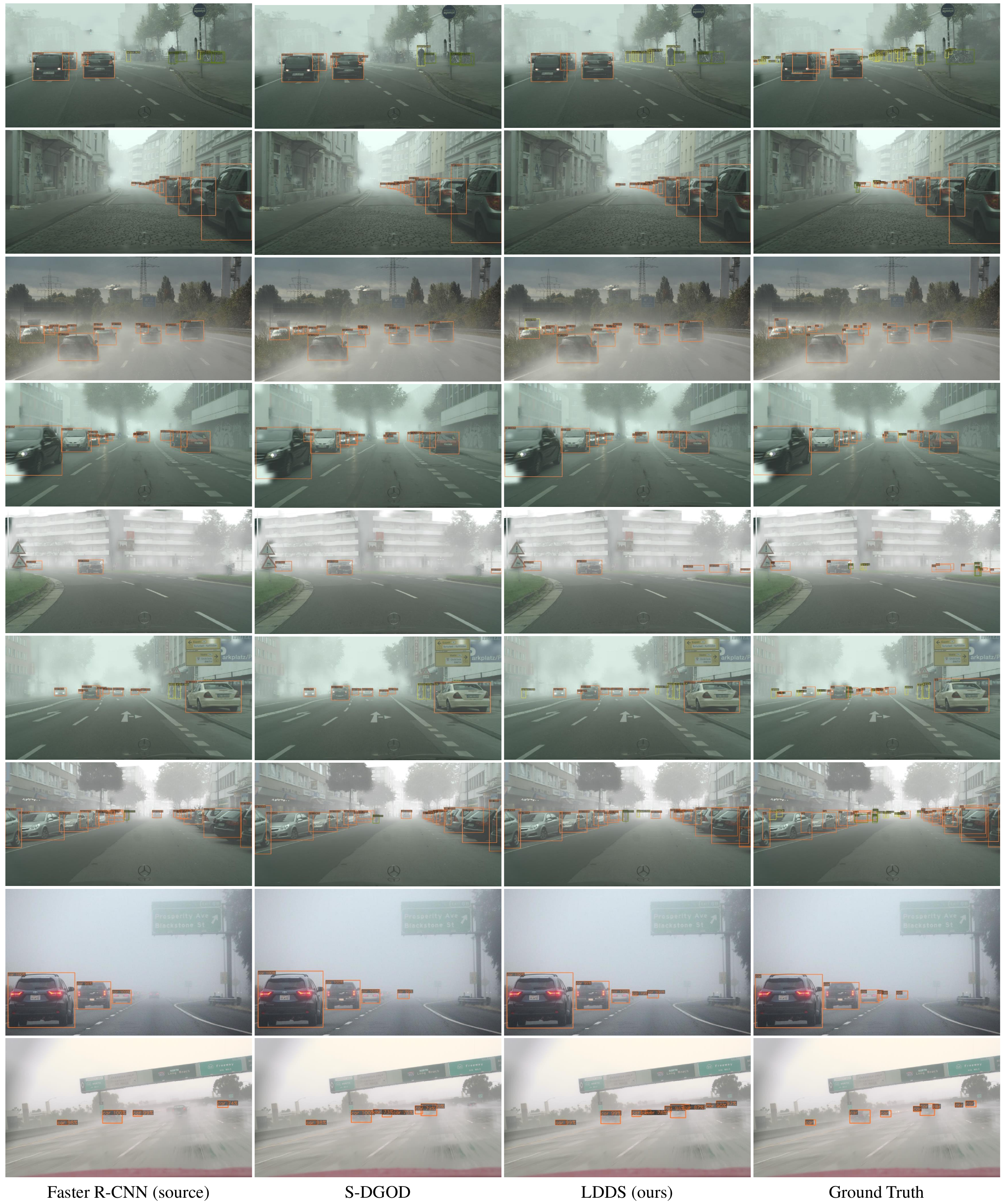}}
\caption{Additional qualitative results on the target domain of daytime-sunny to daytime-foggy for Faster R-CNN (source)~\cite{ren2015faster}, S-DGOD~\cite{wu2022single}, LDDS (ours), and Ground Truth.}
\label{fig:foggy}
\end{figure*}

We have provided additional high-resolution qualitative comparison results. \cref{fig:night} illustrates the qualitative results for the daytime-sunny to night-rainy task using YOLOv8~\cite{yolov8} as the baseline, comparing the source, LDDS, and ground truth. \cref{fig:foggy} shows the results for the daytime-sunny to daytime-foggy task with Faster R-CNN~\cite{ren2015faster} as the baseline, comparing the source, S-DGOD~\cite{wu2022single}, LDDS, and ground truth.

\section{Discussion}
\label{sec:Discussion}
\subsection{Structure of the backbone} 
The backbone network serving as the feature extraction module is a critical component of the detector. Its structure depends not only on the parameter size but also on the architecture type of the detector. For two-stage detectors~\cite{ren2015faster,he2017mask}, the backbone is specialized in feature extraction to the region proposal, which requires a more complex structure to maintain accuracy. On the other hand, the backbone of a one-stage detector is coupled with the detection head~\cite{redmon2016you,redmon2017yolo9000,yolov8}, enabling it to be designed in a more lightweight manner, thereby significantly improving inference efficiency. In transformer-based detectors, the backbone network is responsible for image feature extraction. The flexibility in backbone selection is critical for optimizing the trade-off between speed and accuracy. Therefore,  our goal is to design an SDG technique applicable free from the constraints of the backbone. Current SDG object detection methods based on VLMs require the backbone structure to match that of the image encoder~\cite{fahes2023poda,vidit2023clip}, thereby restricting applicability across different detector models. Our LDDS eliminates this requirement while utilizing the semantic information of VLMs. Furthermore, the demonstrated capability of VLMs in improving model generalization, suggests that our LDDS could be extended to other computer vision tasks as well.

\subsection{Limitation}
While our LDDS outperforms state-of-the-art methods on several metrics, there is still scope for improvement in specific unseen domains. As illustrated in~\cref{fig:night_limit}, using YOLOv8 as the baseline detector in the night-rainy dataset led to missed detections of poorly illuminated vehicles and small blurry objects.
One contributing factor is the dataset itself, where the camera faces direct exposure to high-beam headlights, leading to severe overexposure and a loss of object details. Another factor is that the performance of the baseline detector is inherently constrained and quickly reaches its limit, even with the deployment of SDG methods. To mitigate these issues, besides constructing a more refined dataset, a potential solution would be to replace it with a larger model such as the YOLOv8-L, or to use the more advanced detector YOLOv11~\cite{yolov8} directly. Another feasible strategy is to construct more specialized style prompts to generate richer semantic information that can support domain generalization more effectively in specific unseen domains.

\end{document}